\documentclass[conference]{IEEEtran}
\IEEEoverridecommandlockouts

\usepackage{cite}
\usepackage{amsmath,amssymb,amsfonts}
\usepackage{algorithmic}
\usepackage{graphicx}
\usepackage{textcomp}
\usepackage{xcolor}

\def\BibTeX{{\rm B\kern-.05em{\sc i\kern-.025em b}\kern-.08em
    T\kern-.1667em\lower.7ex\hbox{E}\kern-.125emX}}
\begin{document}

\title{mmID: High-Resolution mmWave Imaging for Human Identification}

\author{\IEEEauthorblockN{Sakila S. Jayaweera, Sai Deepika Regani, Yuqian Hu, Beibei Wang, and K. J. Ray Liu}
\IEEEauthorblockA{Origin Research, Rockville, MD 20852, USA. }
\IEEEauthorblockA{University of Maryland, College Park, MD 20742, USA.} 
}

\maketitle

\begin{abstract}

Achieving accurate human identification through RF imaging has been a persistent challenge, primarily attributed to the limited aperture size and its consequent impact on imaging resolution.
The existing imaging solution enables tasks such as pose estimation, activity recognition, and human tracking based on deep neural networks by estimating skeleton joints. In contrast to estimating joints, this paper proposes to improve imaging resolution by estimating the human figure as a whole using conditional generative adversarial networks (cGAN). In order to reduce training complexity, we use an estimated spatial spectrum using the MUltiple SIgnal Classification (MUSIC) algorithm as input to the cGAN. Our system generates environmentally independent, high-resolution images that can extract unique physical features useful for human identification. We use a simple convolution layers-based classification network to obtain the final identification result. 
From the experimental results, we show that resolution of the image produced by our trained generator is high enough to enable human identification. Our finding indicates high-resolution accuracy with $5\%$ mean silhouette difference to the Kinect device.  Extensive experiments in different environments on multiple testers demonstrate that our system can achieve $93\%$ overall test accuracy in unseen environments for static human target identification.
\end{abstract}
 

\section{Introduction}
\label{sec:introduction}

With the advancement of WiFi sensing technology\cite{liu_wang_2019, 8350392, wu2022wi}, using WiFi reflections for imaging has become an increasingly popular area of research. While conventional imaging techniques based on camera and depth sensors often face issues such as privacy invasion, interference from lighting, and obstruction, WiFi-based imaging offers a solution that preserves privacy while still providing accurate human sensing, even in poor lighting conditions. This opens up new possibilities for applications such as pose estimation\cite{zhao2019through, zhao2018rf, jiang2020towards}, activity detection\cite{10.1145/3349624.3356768}, and human tracking\cite{8804831, 9155293, 10.1145/3498361.3538926}, etc. However, human identification using WiFi-based imaging remains a challenge due to the limited resolution caused by the large wavelength, narrow bandwidth, and the limited number of antennas\cite{Wision, xu2017radio, driverauth}. Recent deep neural network-based methods \cite{jiang2020towards, geng2022densepose, 9008282}  show the feasibility of using 2.4GHz WiFi for human pose reconstruction. Still, these methods require advanced data collection setups or are over-fitted to the trained setup and cannot be deployed in untrained environments. 

Besides 2.4/5GHz WiFi, FMCW radars are widely used in human pose reconstruction because the antenna arrays can provide accurate spatial information about distance and angle. Some prior works \cite{zhao2019through, zhao2018rf} rely on specialized FMCW radar design to improve the resolution and reconstruct the human skeletons, while others \cite{mmmesh, sengupta2020mm, 9723439, SHI2022100228} detect pre-defined key points of the human body and use deep neural networks for 2D/3D pose reconstruction. Even though the above methods achieve high performance for dynamic human reconstruction, they do not focus on extracting features for identification (ID). Most mmWave-based human ID approaches \cite{9440988, ZHAO2021102475} rely on the gait features, which need people to walk and may not be accurate when the gait patterns change. Imaging-based human ID could achieve more robust performance, but there are few prior studies using mmWave imaging due to the limited resolution. 

Recently, the improved resolution of mmWave sensing has enabled new applications, including handwriting tracking \cite{regani2021mmwrite}, keystroke recognition \cite{mmkey2021}, vital sign monitooring\cite{vimo}, super-resolution imaging \cite{mmEye}, and sound sensing \cite{RadioMic, RadioSes}. 
Motivated by those designs, in this paper, we propose a mmWave imaging-based human ID system (\textit{mmID}) with a conditional generative adversarial network (cGAN)\cite{Hawkeye} architecture that can
generate higher-resolution images. Specifically, instead of using raw channel impulse response (CIR) as input to the network, we leverage the super-resolution spatial spectrum estimation method \cite{mmEye}, which extracts spatial images by the MUltiple SIgnal Classification (MUSIC) algorithm followed by joint transmitter smoothing.  The image reconstruction network takes the estimated spectrum as input and further improves the resolution by training with the Kinect Depth map as ground truth, whose network architecture consists of an encoder-decoder-based Generator and a 3D and 2D encoder-based Discriminator. After enhancing the resolution, we utilize a convolution layer-based network design to generate the final identification result.

We implemented \textit{mmID} using a standard 60GHz networking chipset with 32 co-located Tx and Rx arrays and conducted experiments in both office and home environments with seven participants in six basic poses. The experimental results demonstrate a high reconstruction accuracy with $5\%$ median silhouette difference to the Kinect depth map. Furthermore, \textit{mmID} achieves a $93\%$ overall test accuracy for static human target identification.

In summary, the main contributions of the paper are as follows. 

\begin{itemize} 
    \item We proposed \textit{mmID} system, which reconstructs the entire human body and can capture the physical differences between individuals that are not extracted by previous methods. 
    \item We implemented \textit{mmID} on a 60GHz chipset and validated the performance through extensive experiments, demonstrating a $93\%$ identification accuracy and showing the feasibility of deploying the proposed design in an unseen environment in real-world scenarios.
\end{itemize}

The rest of the paper is organized as follows. Section \ref{signal_model} introduces the mmWave radar signal model. Section \ref{high_resolution} explains the details of the proposed method. The implementation details and experimental results are presented in Section \ref{results}. Section \ref{conclusion} concludes the paper.

\section{mmWave Radar Signal Model}\label{signal_model}

In this work, we use a 60GHz commodity WiFi device with Qualcomm 802.11ad chipset. An additional array is attached to the chipset to enable radar operation. Each transmitter (Tx) transmits radar pulses and the pulses are received sequentially by the receiver (Rx) antenna array after reflections from the static and dynamic objects. The received pulses are correlated at the Rx side to estimate the CIR. 

Assume that the Tx and Rx antenna arrays have $M$ and $N$ elements, respectively, and the estimated CIR $h_{m,n}(\tau,t)$ between Tx antenna $m$ and Rx antenna $n$ can be written as

\begin{align}
    h_{m,n}(\tau,t) = \sum_{k=0}^{K-1} a^k_{m,n}(t) \delta(\tau - \tau_k(t)),
\end{align}
where $K$ is the number of CIR taps, $\delta$ is the delta function, $a^k_{m,n}(t)$ and $\tau_k(t)$ is the complex amplitude and propagation delay of the $k^{th}$ tap, respectively. At a given time $t$, the 60GHz device captures $M \times N \times K$ complex values as a 3D matrix. 

The 60GHz device has co-located Tx and Rx antenna arrays with 32 elements each. Due to the high bandwidth of 3.52GHz, the device offers $\Delta \tau = 0.28$ ns time resolution, or a $4.26$ cm range resolution \cite{mmEye, regani2021mmwrite} providing a unique advantage over 2.4/5GHz WiFi devices with short bandwidths. 
In addition, the high carrier frequency of 60GHz is more resistant to the multipath effect and can achieve high imaging resolution with a smaller aperture size.

\begin{figure}[t]
\centerline{\includegraphics[width = 3.5in]{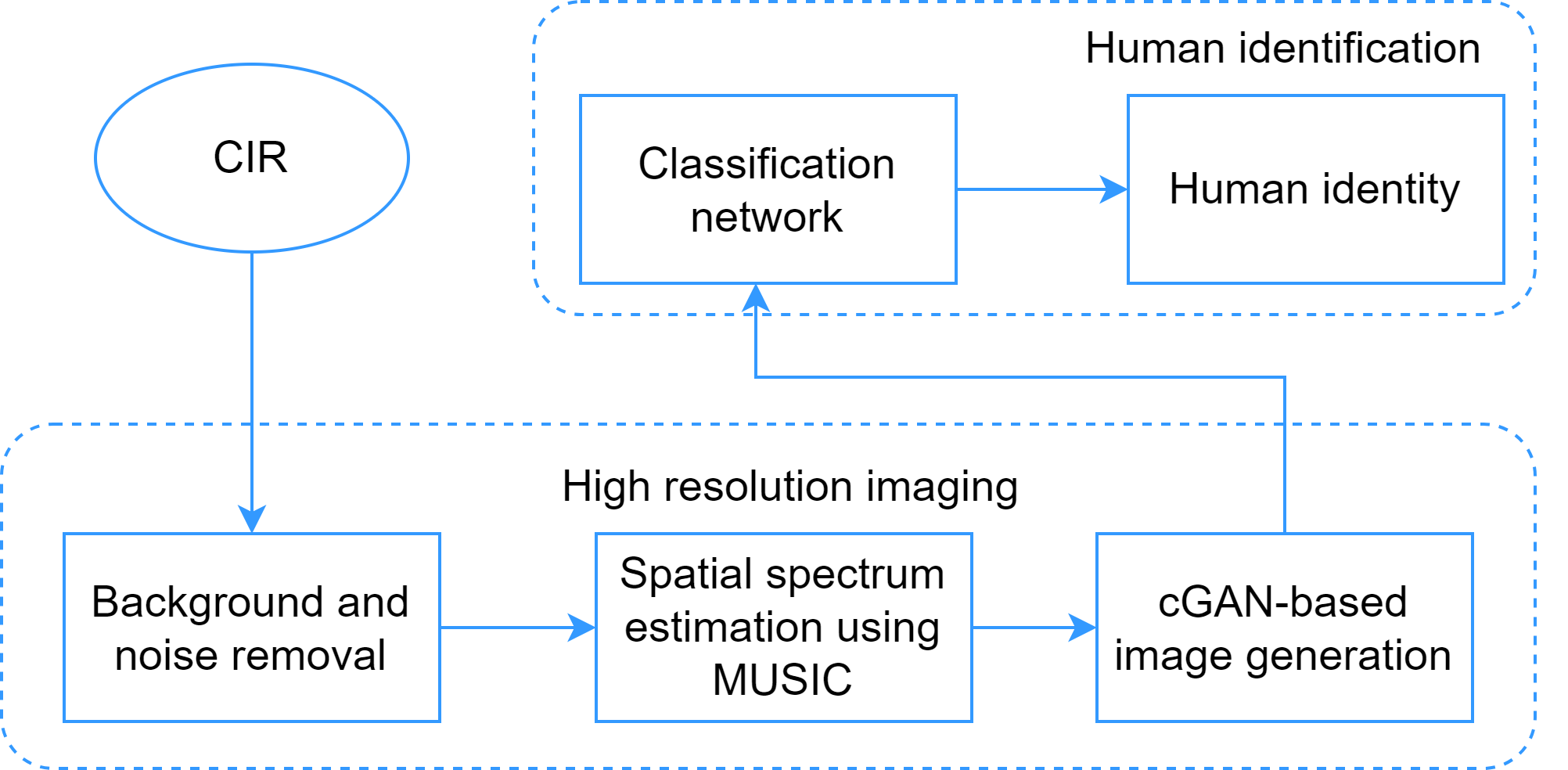}}
\caption{mmID System Design}
\label{fig:system}
\end{figure}

\section{mmID Design}\label{high_resolution}
This section presents the details of \textit{mmID} design, which consists of two key modules as shown in Fig. \ref{fig:system}. First, we achieve high-resolution imaging of the whole human body leveraging a cGAN, and then we use the generated image to identify a human using a CNN-based neural network.

\subsection{Background Removal}
Here, we aim to capture the static human body with high resolution. However, besides the target human, the transmitted pulses are also reflected from background objects, resulting in low imaging performance and environmentally dependent system design. Moreover, internal signal reflections on the Intermediate frequency (IF) cable connectors can further affect the imaging quality.  Thus, as illustrated in Fig. \ref{fig:system}, we first eliminate the background information and noise before processing. To precisely remove the background, we collect CIR $h_{m,n}^{\text{e}}$ for a few seconds without any targeted human. Assuming that we collect $S$ samples of empty data, the noise and background removed CIR can be obtained as 
\begin{align}
    h_{m,n}^c(\tau,t) = h_{m,n}(\tau, t) - \alpha \times \frac{1}{S} \sum_s h_{m,n}^e(\tau, t_s),
\end{align}
where $\alpha$ can be calculated to minimize the energy of the first $K_0$ taps. We choose the first $K_0$ taps, which are not affected by the presence of a user and are only affected by internal noise. This leads to more accurate background and noise removal \cite{mmEye}.

\subsection{Spatial Spectrum Estimation with MUSIC}
Due to the limited spatial resolution of conventional beamforming, we utilize the MUSIC algorithm \cite{MUSIC} for better imaging quality. Here we define $4\times4$ steering subarrays $S$ from the $6\times6$ receiver array. Steering vector $S$ includes the phase response of the antenna array from the direction $(\theta, \phi)$. Then, CIR $h$ can be formulated as
\begin{equation}
    h = Sx + \epsilon,
\end{equation}
where $x$ represents the complex amplitude of the received signal and $\epsilon$ denotes the additive noise. Then, the correlation matrix $R$ can be calculated as
\begin{align}
    R &= \mathbb{E}[hh^H] \nonumber\\
    &= Sxx^HS^H + \mathbb{E}[\epsilon \epsilon^H ] \nonumber \\
    &= R_s + R_\epsilon.
\end{align}
The spatial spectrum for any direction $(\theta, \phi)$ can be obtained as
\begin{align}
    P(\theta, \phi ) = \frac{1}{S^H \mathcal{V}_\epsilon \mathcal{V}_\epsilon^H S},
\end{align}
where $\mathcal{V}_\epsilon$ is the noise subspace that is formed by eigenvectors associated with the $M$ smallest eigenvalues of $R$, and $M$ is the order of $R_s$. Large values of spatial spectrum $P(\theta, \phi)$ indicate the presence of reflected signals.

We use spatial and temporal smoothing to address the rank deficiency issue when using MUSIC. Joint transmitter smoothing (JTS) \cite{mmEye} is carried out to boost the resolution further.

\subsection{Human Body Reconstruction Design}

\begin{figure}[t]
\centerline{\includegraphics[width = 3.5in]{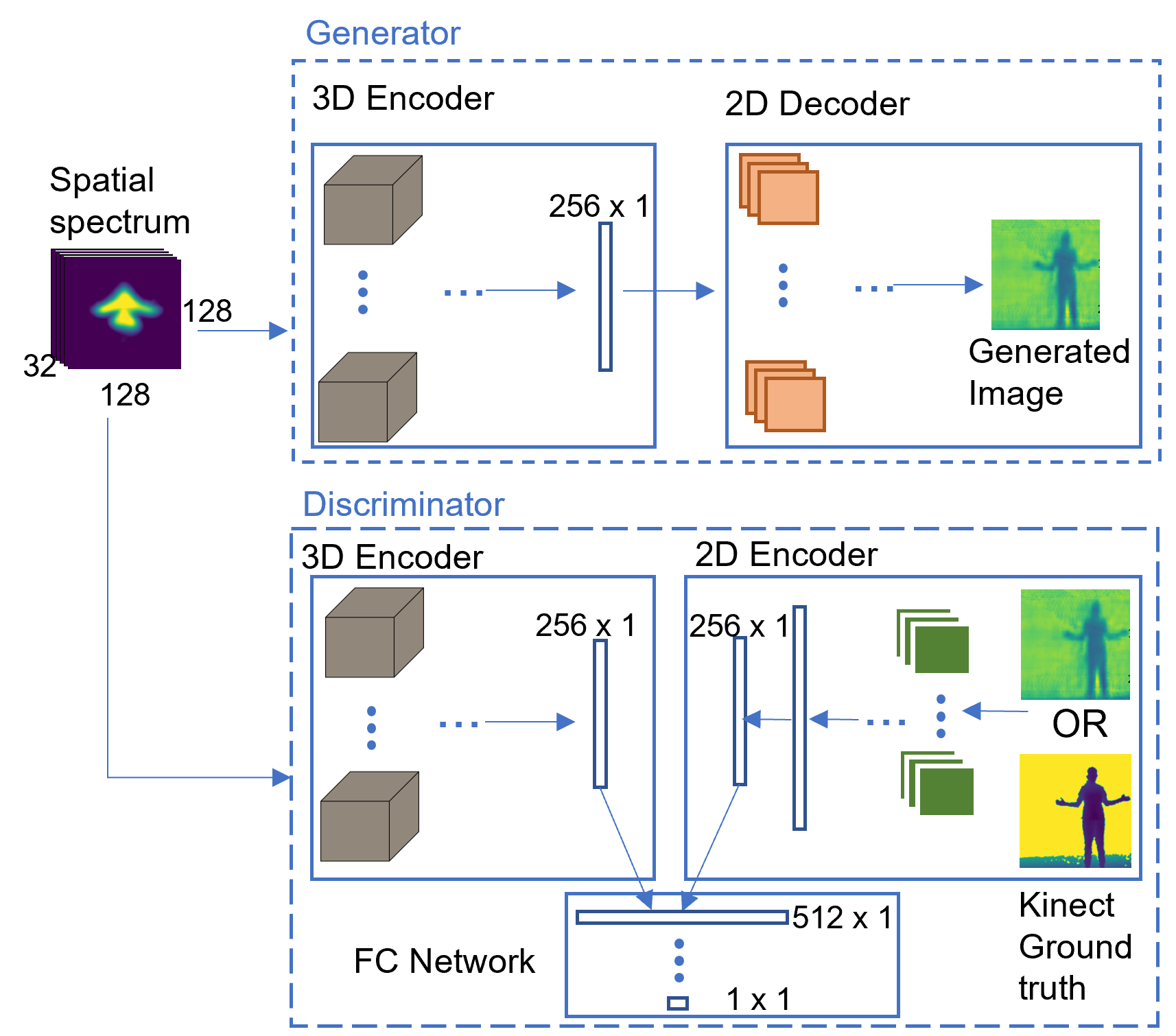}}
\caption{Human Body Imaging Reconstruction Network}
\label{fig:network model}
\end{figure}

The images generated from the MUSIC algorithm have a low resolution and cannot be directly used for human identification as they only contain coarse information and do not include high-resolution details. Thus, we propose to use a cGAN-based architecture as illustrated in Fig. \ref{fig:network model}. The proposed network architecture is based on \cite{Hawkeye}, and we have made several modifications to adapt it to human body imaging. First, while \cite{Hawkeye} uses pre-voxel energy heatmaps as the input, we use the estimated spatial spectrum depth maps, which contain rich spatial information. Second, we additionally use SSIM loss to extract finer structural details of the body to align the generated image output with the ground truth. 
Third, we modify the L1 loss to a weighted L1 loss to preserve important details and improve perceptual quality.

\textbf{Input of the Network:} Our experimental device consists of 32 Tx and 32 Rx antennas. As we perform JTS explained above, we consider a full Rx array when generating one spectrum image. Using a radar frame transmitted from 1 Tx antenna, we estimate a spectrum image of size $128\times128$. Using the full Tx antenna array instead of one antenna allows us to capture high-resolution information. Hence, our input to the network is a 3D spatial spectrum with the size of $128\times128\times32$.

\textbf{Ground Truth:} We use Microsoft Kinect v1 to capture the ground truth depth map. Before training the network using the ground truth data, the background details of the Kinect depth map are removed. The ground truth data has a size of $256 \times 256$.

\textbf{Generator:} The design consists of an encoder-decoder architecture. Since the input is 3-dimensional, the encoder consists of 3D convolution layers followed by batch normalization. The decoder has 2D de-convolution layers, and the generated image is the same size as the Kinect depth map. To address the vanishing gradient issue of the generator design, we used two skip connection layers between the 3D encoder and decoder layers\cite{Hawkeye}.

\textbf{Discriminator:} The network has two stream architecture with 3D encoder and 2D encoder. The 3D encoder is similar to the generator encoder and uses the same 3D spatial spectrum input. On the other hand, the 2D encoder takes the output image from the generator or Kinect ground truth image. Two output vectors of the size $256\times1$ from each encoder network are concatenated at the end and fed into the fully connected (FC) network to the final voting. The FC network classifies the input as real vs. generated to train the discriminator network and improve the Generator network performance based on the Discriminator.

\textbf{Loss Function:} During the training, we use weighted summation of four different loss functions to calculate the final loss. Unlike our previous work \cite{} here we use two modalities when calculating the loss function. Similar to \cite{Hawkeye}, we use GAN loss $L_G$ and perceptual loss $L_p$ using the feature space of the VGG network.  In addition, the L1 loss is replaced by the weighted L1 loss function \cite{8578442} $L_{1-weighted}$. Our output and ground truth images have more background pixels than the target image pixels. Hence, treating all pixels equally penalizes important regions. Thus, we assign lower weights for the background pixels and higher importance for the regions of interest to improve the performance.  Moreover, structural similarity index (SSIM) loss $L_{SSIM}$ is calculated between the generator's output and ground truth. SSIM loss can capture structural information and visualization quality compared to the pixel-wise loss function. Also, SSIM loss is robust to noise and artifacts, which can benefit mmWave-based image generation. Suppose the ground truth is $y$, and the generated image is $\hat{y}$. The loss functions can be given as
\begin{align}
    L_p &= \mathbf{E}||VGG(y) - VGG(\hat{y})||_1\\
    L_{1-weighted} &= \sum_{\forall i} w_i|y_i - \hat{y}_i|\\
    L_{SSIM} &= -SSIM(y, \hat{y})\\
    L_{final} &= L_G + \lambda_1 L_{1-weighted} + \lambda_p L_p + \lambda_s L_{SSIM}.
\end{align}

Here, the SSIM index takes values between -1 to 1, indicating 1 as the perfect similarity. Considering minimizing loss function as the typical objective, we use negative SSIM as a loss. Weighted summation of all loss functions is employed to balance the various aspects of the performance. The values of $\lambda_1, \lambda_p$ and $\lambda_s$ is tuned in grid search manner.

\begin{figure*}[t!]
\begin{minipage}[b]{0.3\linewidth}
  \centering
  \centerline{\includegraphics[width=1.6in]{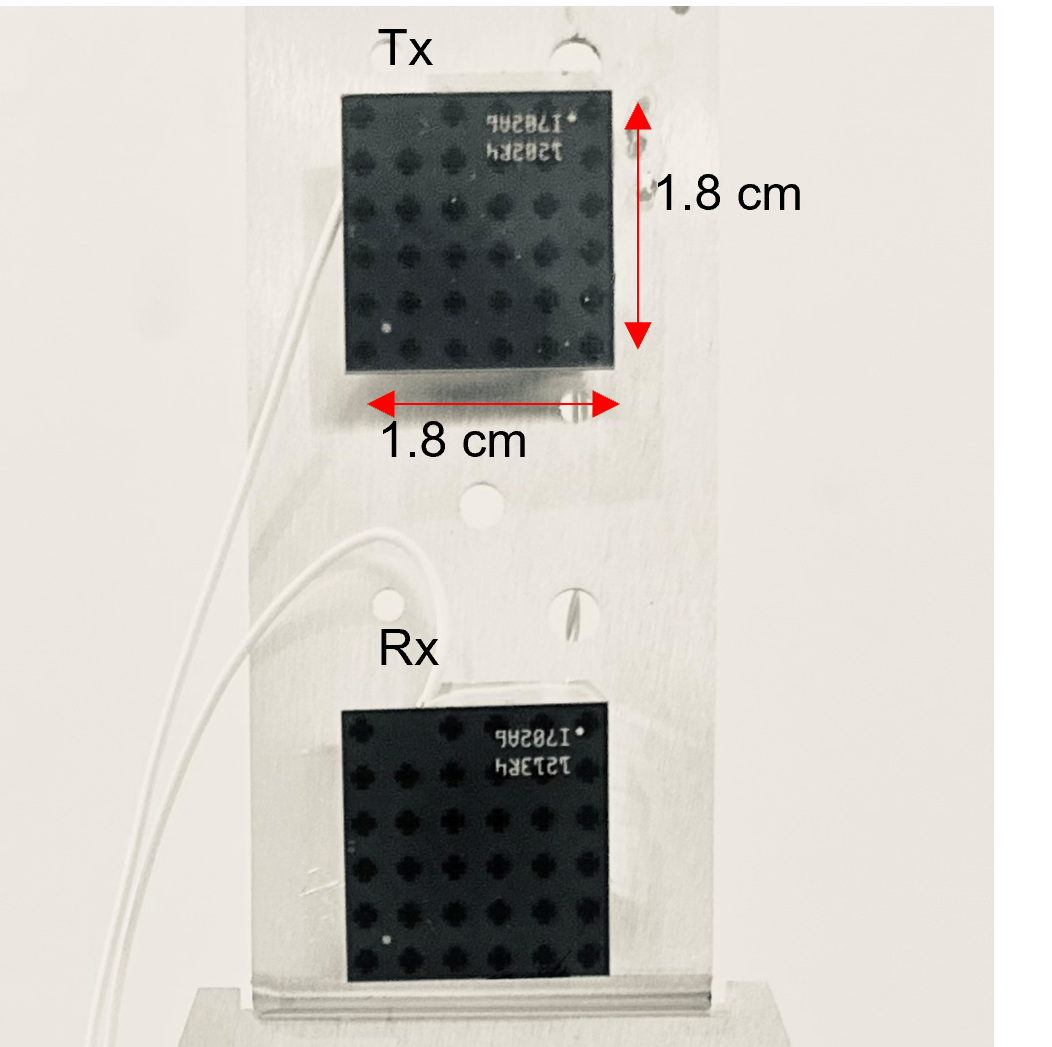}}
  \centerline{(a) }\medskip
\end{minipage}
\begin{minipage}[b]{0.3\linewidth}
  \centering
  \centerline{\includegraphics[width=2.6in]{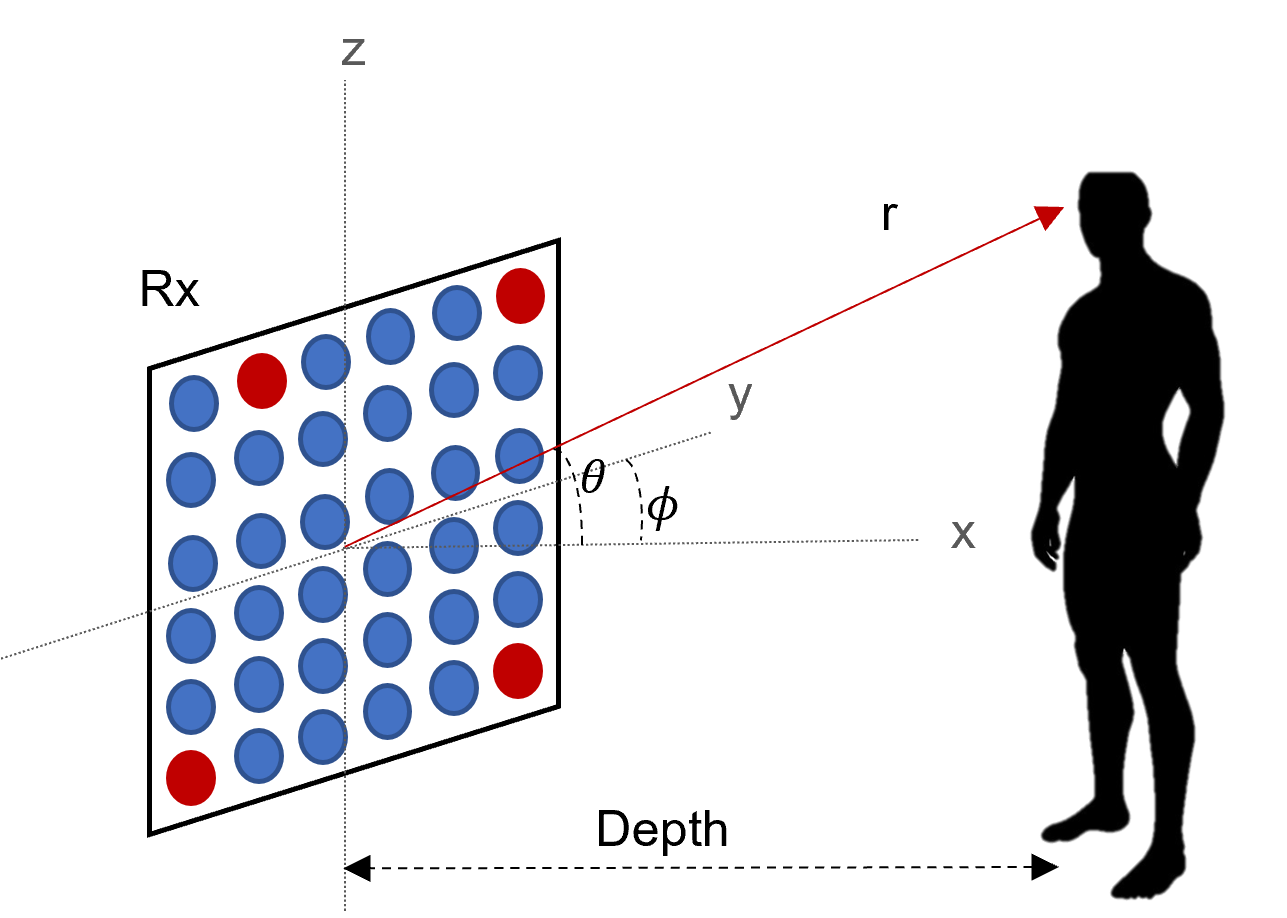}}
  \centerline{(b) }\medskip
\end{minipage}
\begin{minipage}[b]{0.35\linewidth}
  \centering
  \centerline{\includegraphics[width=2.6in]{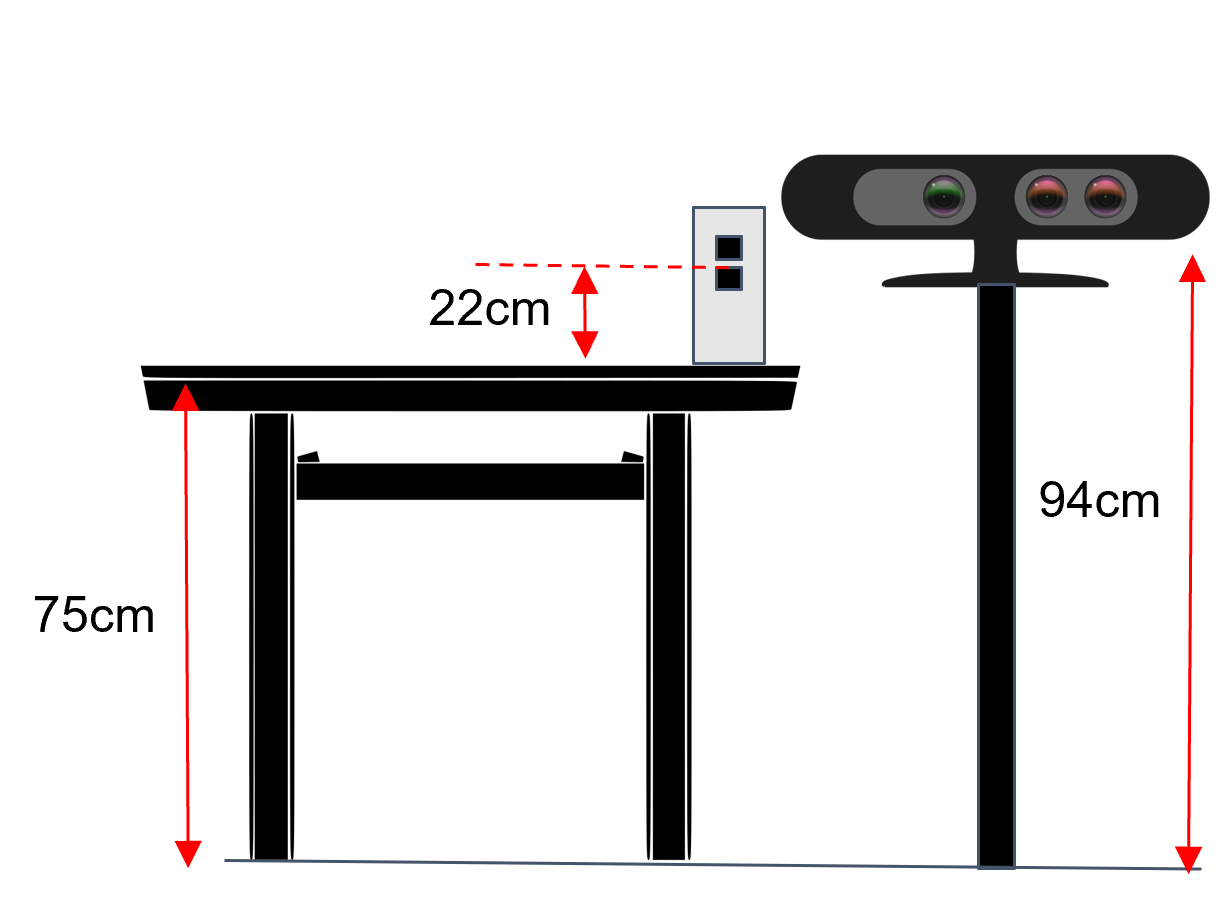}}
  \centerline{(c) }\medskip
\end{minipage}
\caption{Data collection devices and setup. (a)60GHz Qualcomm device with co-located Tx and Rx, (b) $6\times 6$ antenna layout, $\theta$ and $\phi$ denotes the elevation and azimuth angle respectively, (c) Data collection setup with mmWave device and Kinect.}
\label{fig:datacollection}
\end{figure*}
 
\subsection{Imaging-Based Human Identification}\label{classifier}
As the reconstructed images do not have a high enough resolution in human faces, \textit{mmID} is based on the human body shape. This may lead to a certain confusion in identifying two targets with similar body shapes. However, the experiments in later sections show that \textit{mmID} can be used for household-level human identification tasks.

The classifier is based on four convolution layers followed by max-pooling. In the end, two fully connected layers are used to predict the final identity. The last layer has log softmax activation, while all the other layers use ReLU activation. 

\section{Experimental Results}\label{results}
\subsection{Dataset}
There is no publicly available 60GHZ mmWave dataset for human body imaging. Therefore, we built a dataset by collecting data using a commodity 60GHz 802.11ad chipset provided by Qualcomm as shown in Fig. \ref{fig:datacollection}(a).  The co-located Tx and Rx antenna arrays have 32 elements each and are arranged in a $6\times6$ grid with $3$mm separation. The antenna arrangement is depicted in Fig. \ref{fig:datacollection}(b) with 4 missing locations marked in red.  We collect the ground truth depth map using Microsoft Kinect V1 by the ``libfreenect" library. The mmWave device and Kinect are placed side-by-side to align the data, as illustrated in Fig. \ref{fig:datacollection}(c).

We collected data in three typical indoor environments, including a home and two office spaces, resulting in entirely different sensing environments and reflection backgrounds. The dataset includes seven people in 6 different poses. For each person, we collected around 200 image samples. During data collection, each participant stood at a distance of 1.5-2 meters in front of the mmWave device. We used geometric transformations, including flipping, shifting, and rotation, to further augment the dataset.

\subsection{Training}
The proposed cGAN is trained on an NVIDIA GeForce RTX 4090 GPU. We trained the network for 200 epochs using the ADAM optimizer. In the first 100 epochs, Generator and Discriminator were trained using 0.001 and $1\times10^{-5}$ learning rates, respectively. Then, the learning rate decays exponentially for the following 100 epochs. 

The convolution layers-based classifier was trained using a stochastic gradient descent (SGD) optimizer starting with $1\times10^{-4}$ learning rate. We trained the network using 100 epochs, and the learning rate decayed by a factor of 0.1 after every 10 epochs.

\subsection{Performance}
We test the performance of the proposed system using $20\%$ of data from the dataset. Moreover, we evaluate the system performance for data collected in a new environment to validate environmental independence. 

\begin{figure}[t]
\begin{minipage}[b]{0.32\linewidth}
  \centering
  \centerline{\includegraphics[width=1.1in]{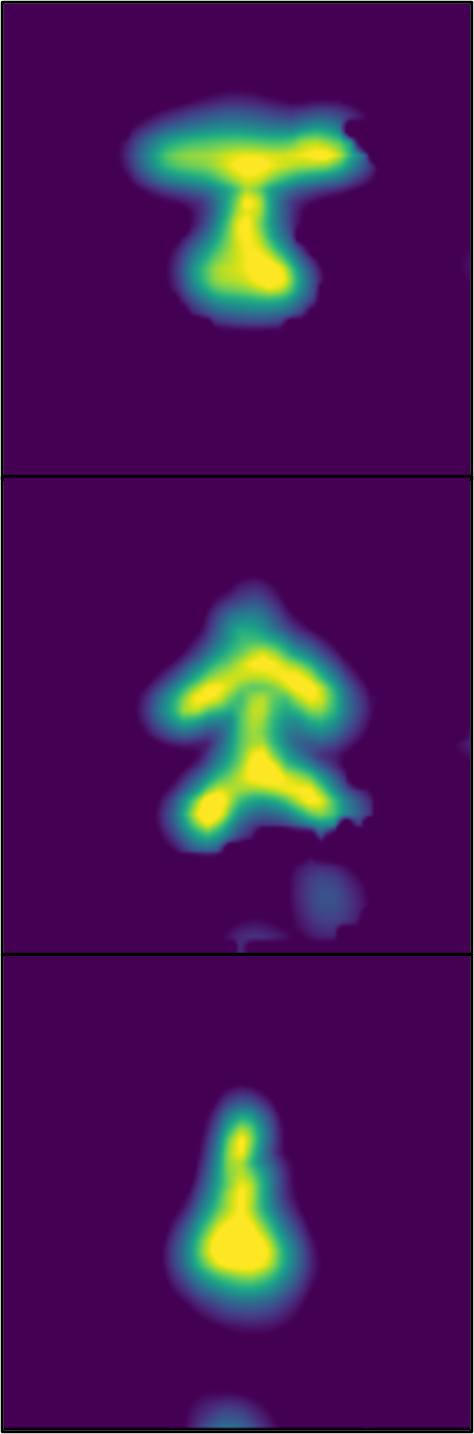}}
  \centerline{(a) }\medskip
\end{minipage}
\begin{minipage}[b]{0.32\linewidth}
  \centering
  \centerline{\includegraphics[width=1.1in]{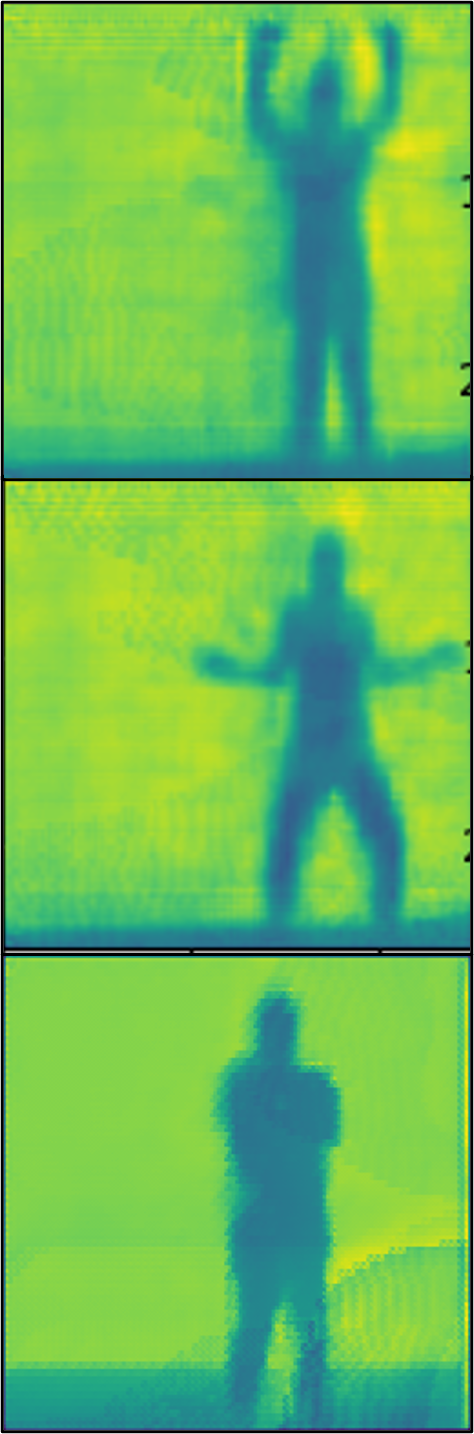}}
  \centerline{(b) }\medskip
\end{minipage}
\begin{minipage}[b]{0.32\linewidth}
  \centering
  \centerline{\includegraphics[width=1.1in]{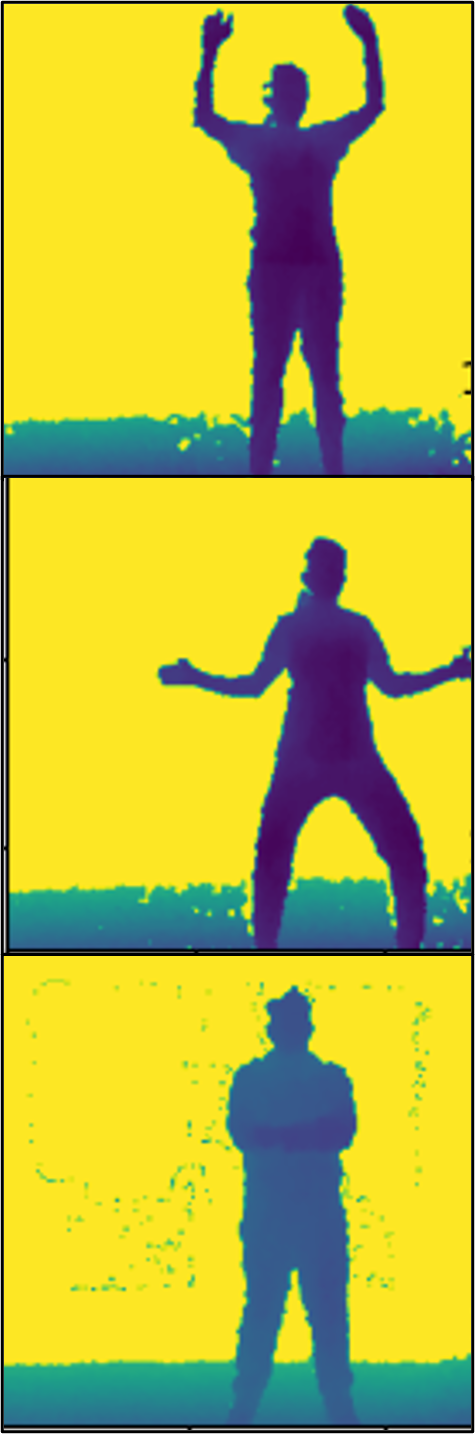}}
  \centerline{(c) }\medskip
\end{minipage}
\caption{Human body imaging performance. (a) mmWave spectrum estimated using MUSIC for one Tx, (b) Generated image, and (c) Ground truth image obtained using Kinect.}
\label{fig:results}
\end{figure}

\textbf{Human Imaging:} Fig. \ref{fig:results} shows human imaging results generated by the trained generator. The estimated spectrum from the 32 Tx antennas is used as the input to the network. For visualization purposes, in Fig. \ref{fig:results}, we show an example spectrum obtained from one transmitter. As shown, the spectrum has low resolution, while we can obtain comparatively higher resolution using a trained generator network.

We evaluate the quality of the generated image using the Silhouette difference (SD) metric similar to \cite{mmEye}.  SD metric calculated the percentage of XOR difference between the generated image and the Kinect ground truth image, taking a value of 0 if the pixel of the human torso matched with the ground truth and taking a value of 1 otherwise. 

In \cite{mmEye}, SD degrades to $32.3\%$ when the target is 1.5m away from the device. In contrast, the experimental evaluations show that cGAN-based image generation can achieve $5\%$ SD even if the human target is 1.5-2 m away from the device. 

\begin{figure}[t]
\centerline{\includegraphics[width = 3.6in]{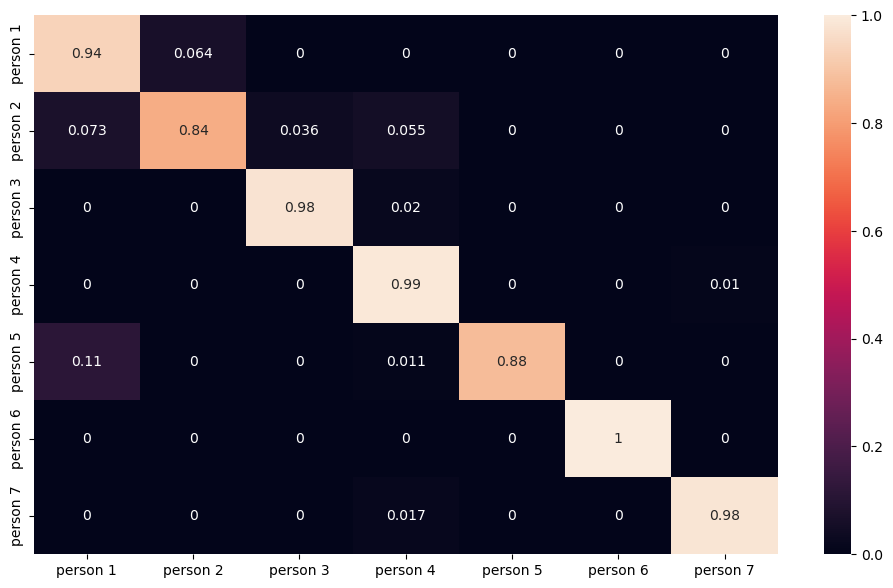}}
\caption{Classification accuracy confusion matrix for unseen environment}
\label{fig:results_classifier}
\end{figure}

\textbf{Identification:} We evaluate the human identification performance under two different setups. First, we evaluate the identification accuracy for the test set obtained from the same environment setup. Then we use a test set obtained from the unseen environment to ensure the system's capability of location independence. Moreover,  the data is collected during different time periods of the day, including night. The overall accuracy for seen and unseen environments are shown in Table \ref{tab:identification}, and we can see that \textit{mmID} can be well generalized for an unseen environment.

\begin{table}[h]
    \centering
    \caption{Identification Accuracy}
    \begin{tabular}{|c|c|c|}
        \hline
        & Test accuracy\\
        \hline
       Seen environment  & $97\%$ \\
       Unseen environment  & $93\%$ \\
       \hline
    \end{tabular}
    \label{tab:identification}
\end{table}

Fig. \ref{fig:results_classifier} shows the resulting confusion matrix for the test data obtained in the unseen environment. It shows that people with similar body shapes, such as persons 1 and 2, can be mixed up with each other and exhibit low identification accuracy. On the other hand, persons 3, 4, 6, and 7 show high accuracy as they have significantly different torso structures.

\subsection{Effects of Loss Function}
\begin{figure}[t]
\begin{minipage}[b]{0.45\linewidth}
  \centering
  \centerline{\includegraphics[width=1.56in]{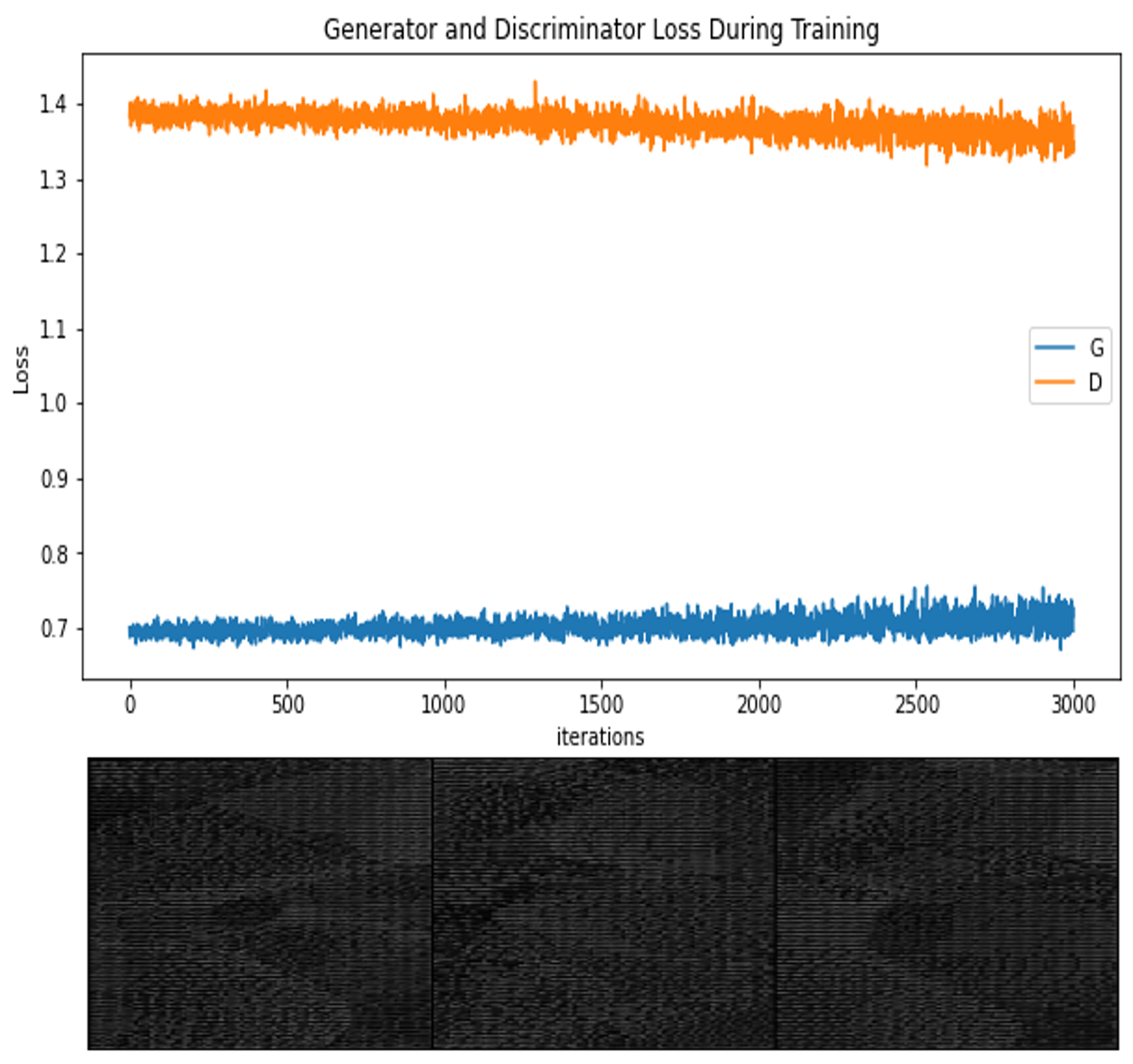}}
  \centerline{(a) }\medskip
\end{minipage}
\begin{minipage}[b]{0.45\linewidth}
  \centering
  \centerline{\includegraphics[width=1.56in]{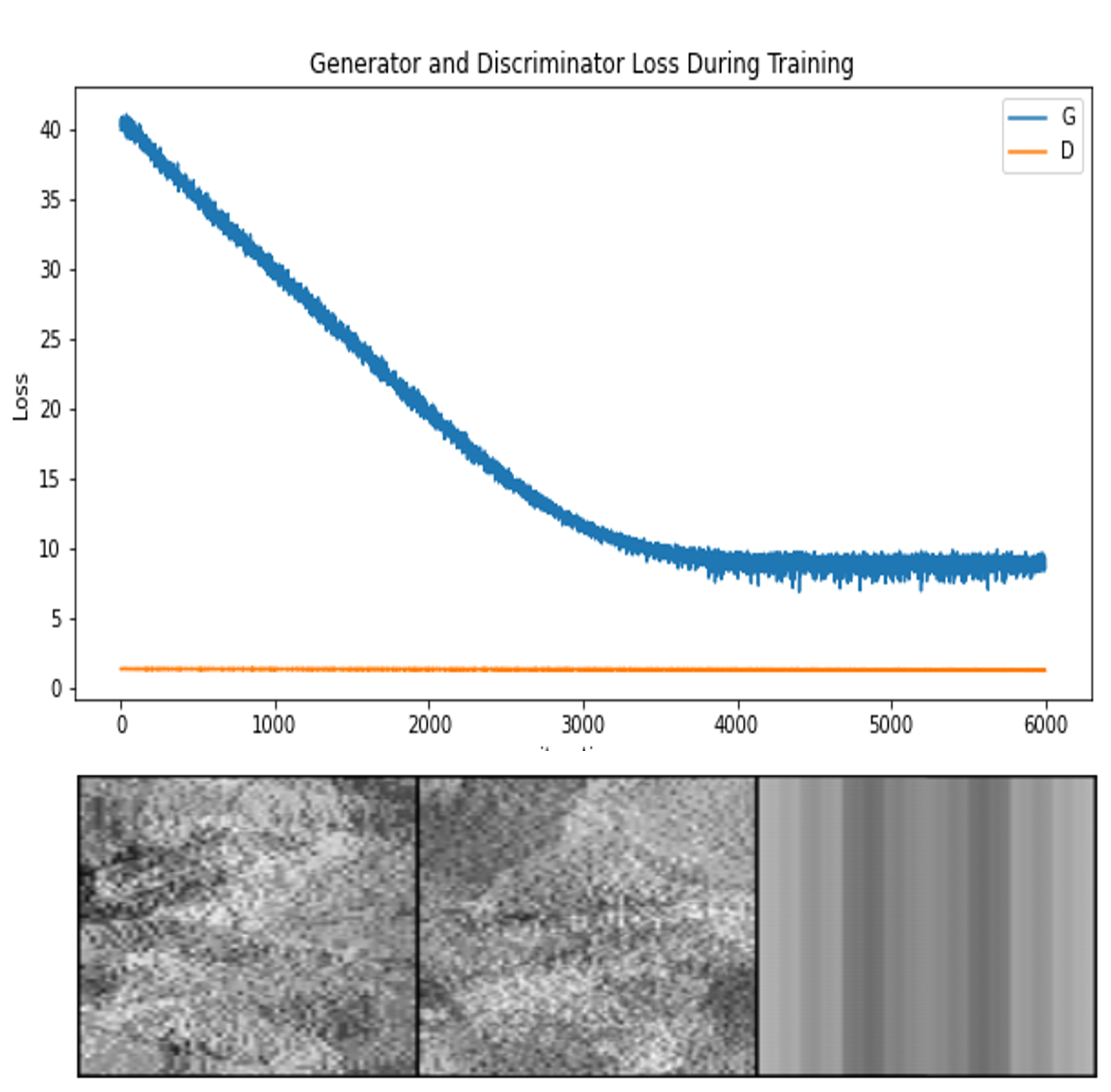}}
  \centerline{(b) }\medskip
\end{minipage}

\begin{minipage}[b]{0.45\linewidth}
  \centering
  \centerline{\includegraphics[height=1.5in, width=1.56in]{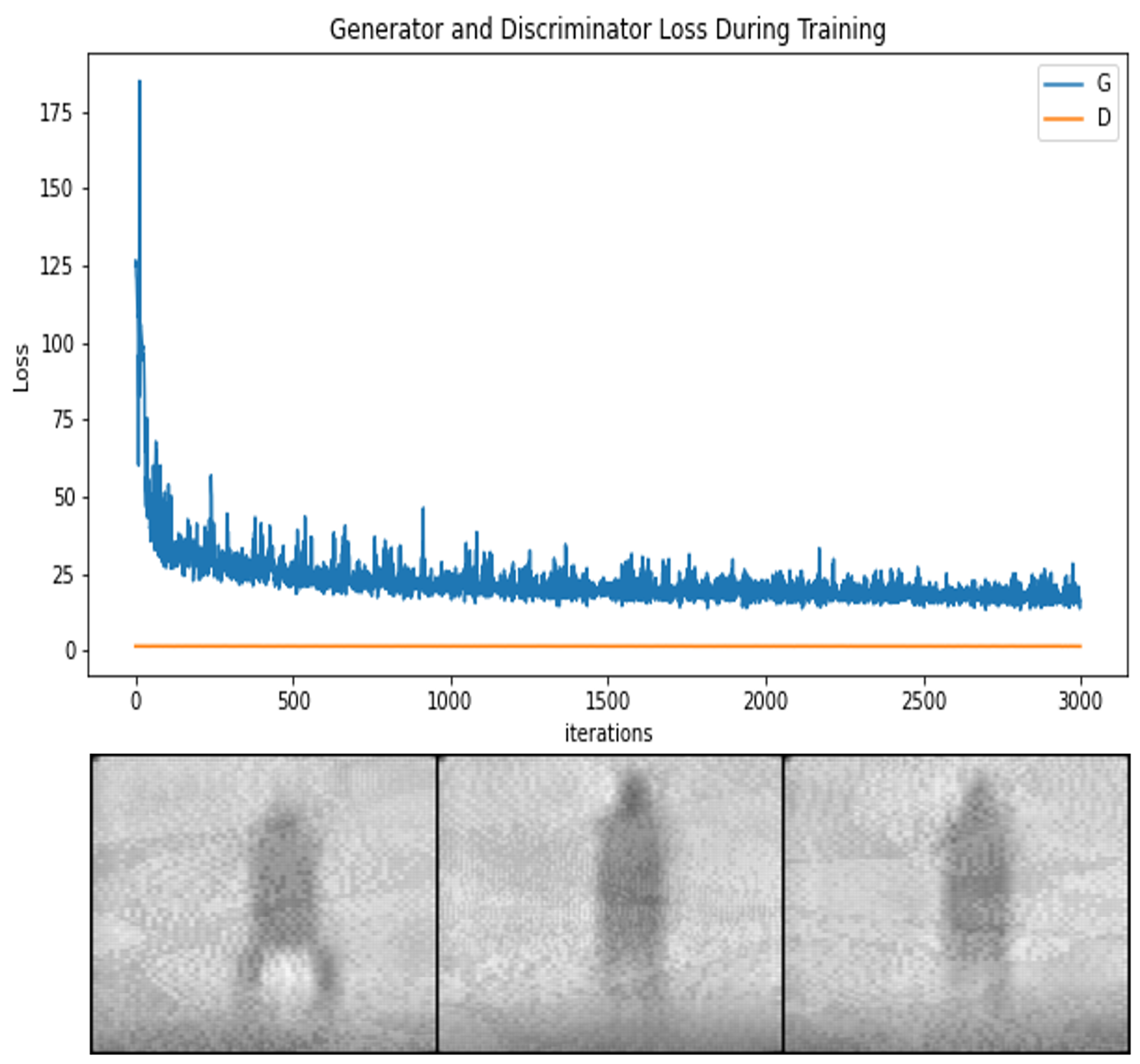}}
  \centerline{(c) }\medskip
\end{minipage}
\begin{minipage}[b]{0.45\linewidth}
  \centering
  \centerline{\includegraphics[height=1.5in, width=1.56in]{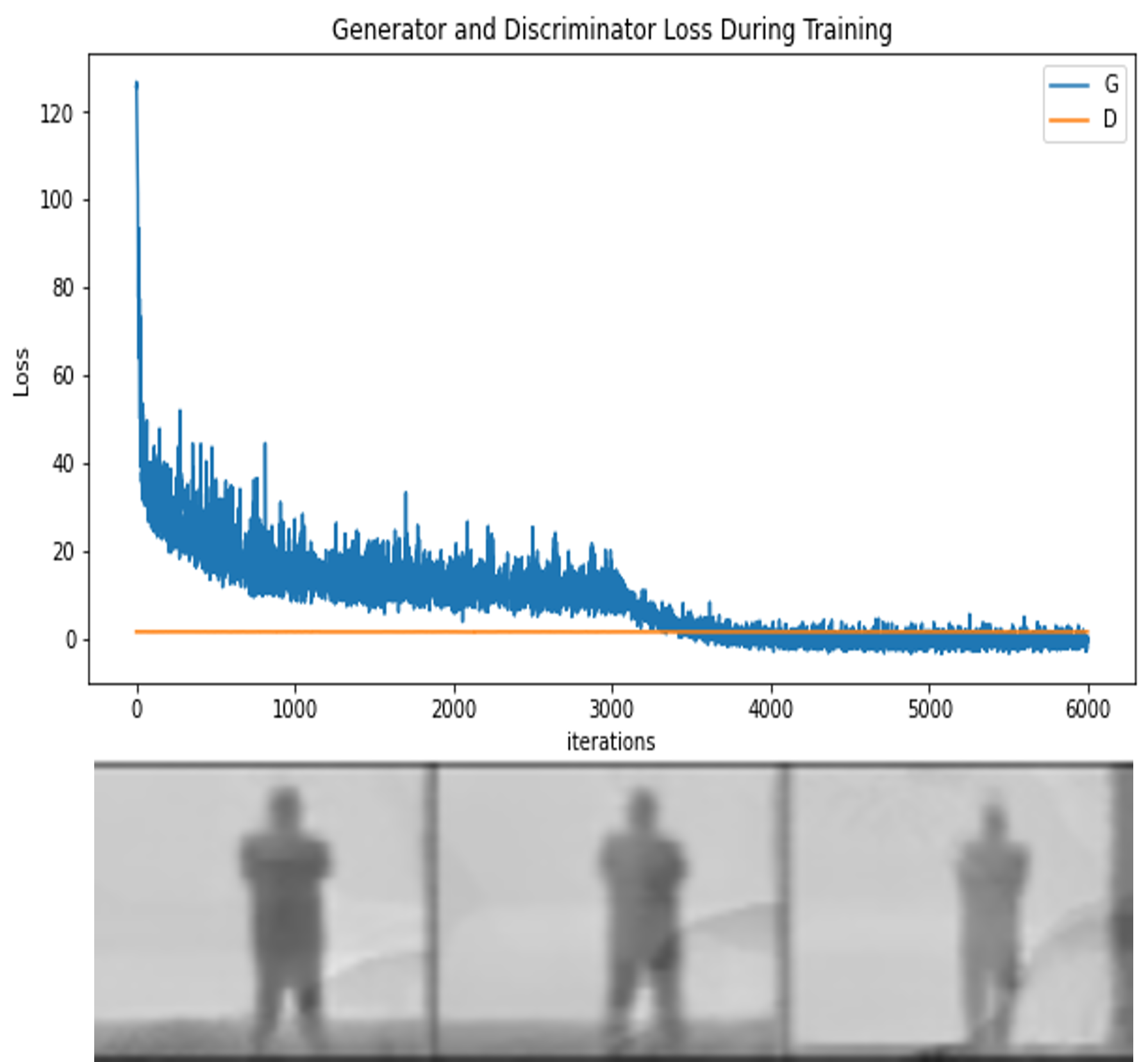}}
  \centerline{(d) }\medskip
\end{minipage}
\caption{Effects of loss functions. (a) Using $L_G$ loss, (b) Using $L_G$ and $L_1$ loss, (c) Using $L_G, L_p$ and $L_1$ and (d) Using $L_G, L_p, L_1$ and $L_{SSIM}$ loss}
\label{fig:loss}
\end{figure}

To evaluate the contribution of each loss function, we trained the cGAN in different loss combinations. Fig. \ref{fig:loss} shows how training is changed when we add each loss to the final calculation. In each figure, the first row depicts the generator and discriminator loss changes with the interactions, and the second row shows the generated image after all iterations. According to the experiment, perceptual loss $L_p$ contributes to extracting the human torso while adding $SSIM$ loss reduces the noise and improves the visualization quality. 

\subsection{Limitations}

\textit{mmID} has several limitations due to the experimental 60GHz testbed. The operating range of the 60GHz device is limited to 10m, and the reflection energy degrades linearly over the range, resulting in low-resolution images. From our experiments, if the target is more than 2m away from the device, imaging resolution is not as expected. Moreover, there is direct leakage of the transmitter signal into the receiver, as well as internal signal reflections due to IF cable connectors creating noisy CIR taps inside the device. This directly affects the imaging quality if the target is nearby. Thus, we had to limit our sensing range to 1m to 2m.

\section{Conclusion}\label{conclusion}
In this paper, we propose a mmWave-based high-resolution imaging system that enables human identification. The imaging system is based on cGAN architecture and learns image reconstruction from the estimated spatial spectrum. Reconstructed images achieve remarkable resolution with only $5\%$ mean silhouette difference between generated images and ground-truth images from Kinect. The Human Identification module learns distinct pose patterns and torso structures with a convolution neural network to identify human targets.  Experiments with seven users show that the proposed system is independent of the location and can achieve $93\%$ overall accuracy. 

\section*{Acknowledgement}
This research is partly supported by the Key Bridge Foundation.

\bibliographystyle{ieeetr}

\bibliography{main_mmID}

\end{document}